%% file: main.tex
\documentclass[11pt]{article}

\input{header}

\title{Grounding Language in Multi-Perspective Referential Communication}

\author{%
Zineng Tang \hspace{1cm} Lingjun Mao \hspace{1cm} Alane Suhr \\
  University of California, Berkeley \\
  \texttt{\{terran, lingjun, suhr\}@berkeley.edu} \\
}

\begin{document}
\maketitle

\input{sections/00_abstract}

\input{sections/01_introduction}

\input{sections/04_task}

\input{sections/05_experiments}

\input{sections/056_human_study}

\input{sections/055_learning}

\input{sections/03_related}

\input{sections/06_discussion}

\bibliography{ref}

\appendix

\input{sections/appendix}

\end{document}

%% file: header.tex
\pdfoutput=1
\usepackage[final]{acl}

\usepackage[table]{xcolor}         
\usepackage{times}
\usepackage{latexsym}

\usepackage[T1]{fontenc}

\usepackage[utf8]{inputenc}

\usepackage{microtype}

\usepackage{inconsolata}

\usepackage{graphicx}

\usepackage[utf8]{inputenc} 
\usepackage[T1]{fontenc}    
\usepackage{hyperref}       
\usepackage{url}            
\usepackage{booktabs}       
\usepackage{amsfonts}       
\usepackage{nicefrac}       
\usepackage{microtype}      
\usepackage{amsmath}
\usepackage{amssymb}
\usepackage{booktabs}
\usepackage{lipsum}
\usepackage{pgfplots}
\usepackage{multirow}
\usepackage{wrapfig}
\usepackage{xcolor}
\usepackage{bbm}

\newcommand\environments{\mathcal{E}}
\newcommand\environment{e}
\newcommand\pose{\rho}

\newcommand\reals{\mathbb{R}}
\newcommand\referents{\mathcal{R}}
\newcommand\referent{r}

\newcommand\targetindex{t}
\newcommand\speaker{s}
\newcommand\listener{l}

\newcommand\refex{x}
\newcommand\refexes{\mathcal{X}}

\newcommand\scene{\mathcal{S}}
\newcommand\observation{o}
\newcommand\observations{\mathcal{O}}

\newcommand\dataset{\mathcal{D}}
\newcommand\referentpolicy{R}
\newcommand{\poses}{\mathrm{P}}
\newcommand\visibleset{\text{Vis}}
\newcommand\fov{\text{Fov}}
\newcommand\coordinates{\mathcal{C}}

\renewcommand{\paragraph}[1]{\noindent \textbf{#1}}

%% file: sections/00_abstract.tex
\begin{abstract}
We introduce a task and dataset for referring expression generation and comprehension in multi-agent embodied environments.
In this task, two agents in a shared scene must take into account one another's visual perspective, which may be different from their own, to both produce and understand references to objects in a scene and the spatial relations between them.
We collect a dataset of 2,970 human-written referring expressions, each paired with human comprehension judgments, and evaluate the performance of automated models as speakers and listeners paired with human partners, finding that model performance in both reference generation and comprehension lags behind that of pairs of human agents.
Finally, we experiment training an open-weight speaker model with evidence of communicative success when paired with a listener, resulting in an improvement from 58.9 to 69.3\% in communicative success and even outperforming the strongest proprietary model.
\end{abstract}

%% file: sections/01_introduction.tex
\section{Introduction} 
\begin{figure}[t!]
\centering
\includegraphics[width=0.999\columnwidth]{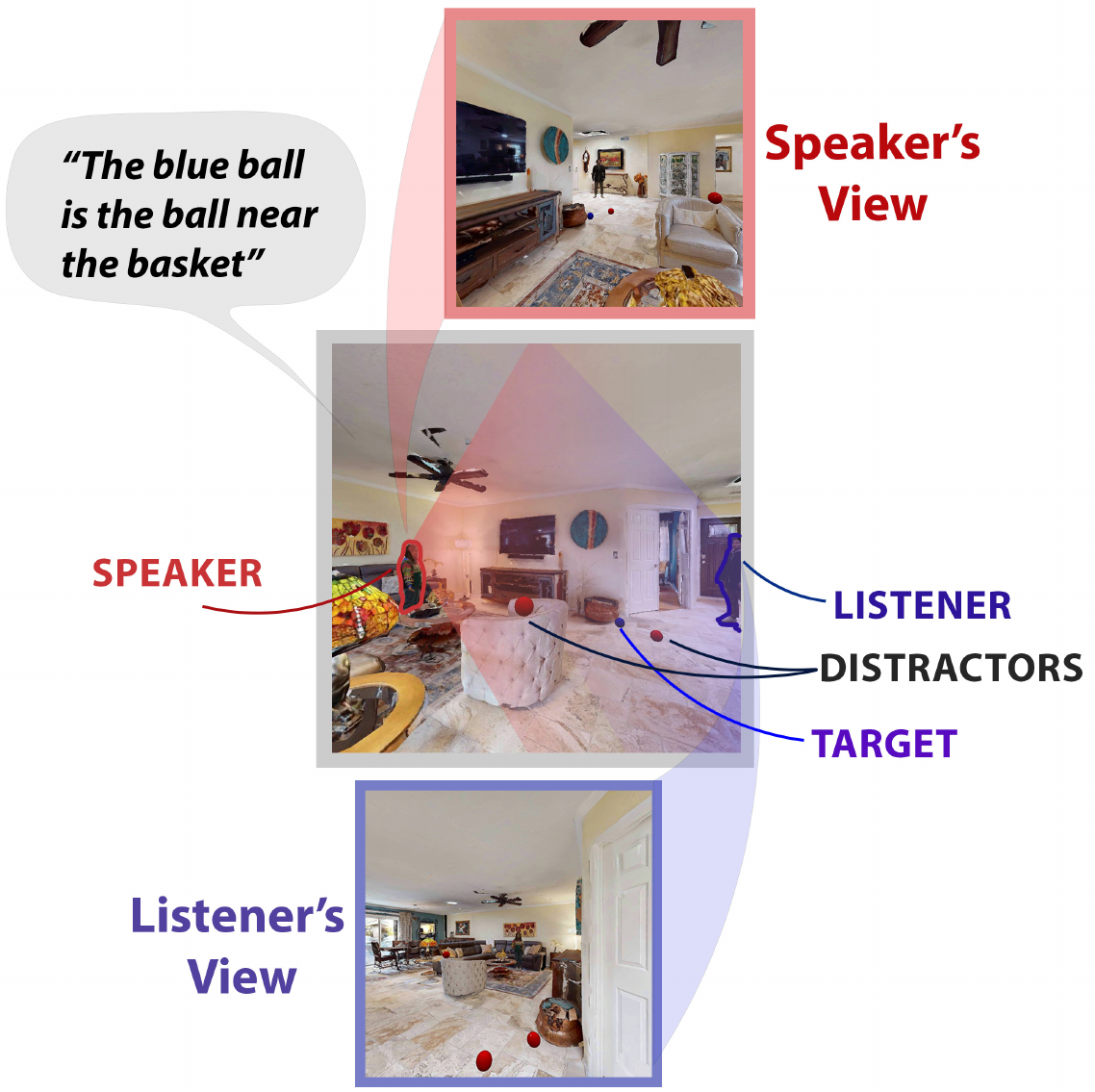}
  \caption{Example scene from our environment and dataset. The center image shows the speaker on the left and the listener on the right with their respective fields of view (FOV).
  The speaker refers to the target object, distinguished by its blue color, and the listener selects the candidate referent they believe is described by the speaker's description, without access to its distinct color. 
  \vspace{-2em}
  }\label{fig:intro}
\vspace{-10px}
\end{figure}

Language agents embodied in situated interactions alongside human users must be able to reason jointly about the space they occupy, the language they encounter, and their human partner's perception. 
For example, consider a home assistant robot that is assisting its human user in finding their lost keys.
This system must take into account its previous and current observations of the space, as well as estimate what the user's current perspective is like in the shared environment.
If the system generates a description of the keys' location that the user clearly and unambiguously understands, they have achieved \textit{communicative success}.
Figure~\ref{fig:intro} shows an example of such a communicative task, where one person describes the location of an object to another person, whose view differs from their own. 
To correctly resolve and generate references to the surrounding environment, both the speaker and listener must take into account the physical relationship between objects, their own view of the environment, and an estimate of the other person's perspective in the environment. 

We study human-human and human-agent referential communication in photorealistic 3D environments, introducing a platform that supports generating task instances with varying levels of difficulty.
In contrast to most prior work on referring expression generation and comprehension, we focus on the setting where both agents are physically embodied in a scene but with different perspectives of the scene.
We collect a dataset of 2,970 human-written referring expressions grounded in 1,485 generated scenes.
We evaluate several recent vision-and-language models on the tasks of referring expression generation and comprehension, including general instruction-tuned vision-language models, models designed for fine-grained vision-language processing, and a modular vision-and-language reasoning system. 
When interpreting human-written referring expressions, the fine-grained Ferret model~\cite{you2023ferret} performs the best, successfully identifying 69.2\% of intended referents.
Using human listeners, we find that the proprietary GPT-4o produces referring expressions that correctly identify the intended target referent for 64.9\% of scenes, while the open-weight LLaVA-1.5~\cite{liu2023improvedllava} is only successful for 55.7\% of scenes.
Compared to the average human-human success rate of 87.6\%, all models lag far behind humans when both generating and comprehending referring expressions.
Analyzing the language used by both automated and human speakers reveals significant differences in referential strategies; for example, human speakers use themselves or the listener agent as reference points much more frequently than automated models, which mostly rely on other objects in the scene.

Our scene-generation platform supports controlling two levels of task difficulty.
First, it supports modifying the relative orientation of the agents.
Second, we train a referent placement policy to minimize communicative success between two automated agents.
For scenes generated using this policy, we see a significant decrease in communicative success across nearly all agent combinations.

Finally, we fine-tune our weaker speaker model, LLaVA-1.5 using data collected during deployment with both human and automated listeners. 
During learning, we first sample referring expressions from the speaker model, convert empirical observations of language interpretation by a listener into training examples~\cite{kojima2021continual}, then apply proximal policy optimization to update model parameters on this data.
We compare our fine-tuned models with GPT-4o, LLaVA-1.5, and human speakers.
With a single round of training and fewer than 200 sampled referring expressions, we see significant improvements in LLaVA-1.5's ability to generate accurate referring expressions, with rates of communicative success with a human listener improving from 58.9 to up to 69.3, outperforming even the originally-stronger GPT-4o speaker. 
This demonstrates the  strengths of learning from interaction to improve communicative success in multi-perspective referential communication.

Our contributions are as follows: 
1. A platform for generating 3D scenes that encompass a two-player referential communication game, enabling the study of multi-perspective referring expression generation and comprehension (Section~\ref{sec:task}). This platform supports controlling task difficulty through the placement of agents and referents.
2. A new dataset of comprising 27,504 sampled scenes, and 2,970 human-written referring expressions grounded in 1,485 generated scenes (Section \ref{sec:data}).
3. A comprehensive analysis of human- and model-written referring expressions, and benchmarking and analysis of different vision and language models on their communicative success (Sections~\ref{sec:eval_settings} and \ref{sec:eval}).
4. An approach for improving an open-source vision-language model on reference generation by learning from communicative success in interaction with human listener agents (Section \ref{sec:commsuccess}).
Our code, models, and dataset are released under an open-source license upon publication at the following URL: \url{https://github.com/zinengtang/MulAgentRef}.

%% file: sections/04_task.tex
\section{Task and Environment}
\label{sec:task}
We study the task of embodied referential communication, where two agents coordinate their attention in a shared scene using referring expressions.
To this end, we design a platform that for generating photorealistic 3D scenes that support this task at varying levels of difficulty.

\subsection{Embodied Referential Communication}
We use a reference game~\cite{clark1986referring}, where a speaker describes a target referent, and a listener attempts to identify the target using the speaker's description.
In our task, two agents are physically embodied in the same shared 3D scene, but with different perspectives, and thus different observations of the scene.
Each scene includes candidate referent objects, one of which is a target object that the speaker needs to communicate to the listener.
Communicative success is achieved if the listener is able to identify the speaker's intended target.

\begin{figure*}[ht]
\centering
\includegraphics[width=0.99\textwidth]{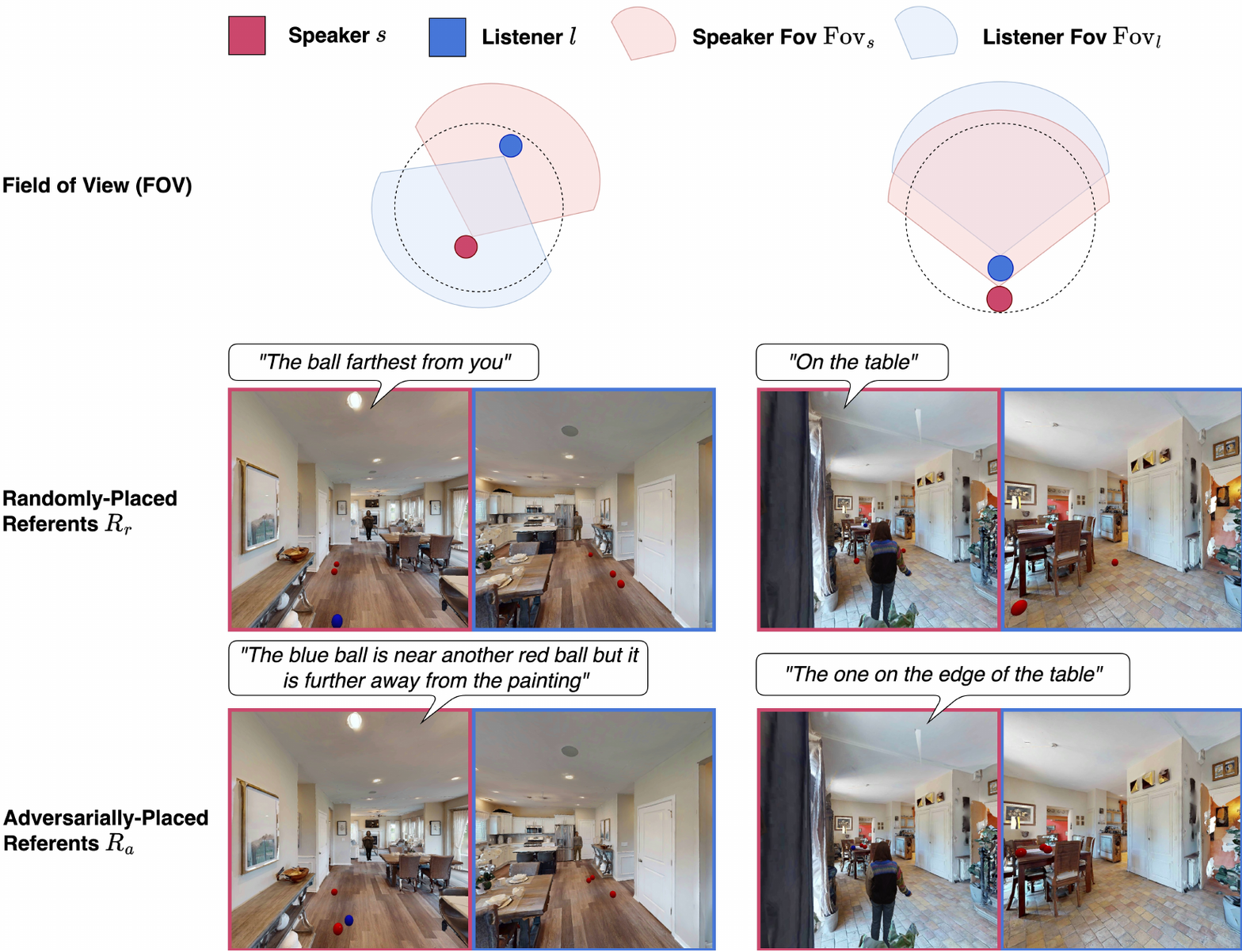}
\caption{Example scenes generated with different relative orientations ($\approx 180^\circ$ on left, $\approx 0^\circ$ on right) and with randomly- (top) or adversarially- (bottom) placed referents.
Adversarially-generated referent configurations often space referents more evenly, with the target referent not easily uniquely identifiable. 
}
\label{fig:view_angle}
\vspace{-12px}
\end{figure*}

Formally, let $\observations$ be the set of possible agent observations, each represented as a 2D image; $\referents$ be the set of candidate referents in an scene, and $\refexes$ be the set of possible referring expressions.
A speaker model $p_{\speaker} : \observations \times \referents^{N} \times \{1 \dots N\} \rightarrow \Delta^{\refexes} $ maps from an observation of the shared scene, a set of referents, and the index of the target referent $\referent_\targetindex$ to a distribution over possible referring expressions.
A listener model $p_{\listener} : \observations \times \referents^{N} \times \refexes \rightarrow \Delta^{\{1 \dots N\}}$ maps from its observation of the scene, the set of all candidate referents, and the referring expression generated by the speaker to a distribution over possible referent indices.
Given a scene with speaker observation $\observation_{\speaker} \in \observations$, listener observation $\observation_{\listener} \in \observations$, a set of $N$ candidate referents $\referents$, and a target referent index $\targetindex$, communicative success is achieved when the listener selects the intended target:

\vspace{-1.5em}
\begin{small}
\begin{align*}
\refex &= \arg \max_{\refex' \in \refexes} p_{\speaker}(\refex' \mid \observation_{\speaker}, \referents, \targetindex) \\
\hat{\targetindex} &= \arg \max_{1 \leq i \leq N} p_{\listener} (i \mid \observation_{\listener}, \referents, \refex) \\
\text{Success}(p_\speaker, p_\listener, \observation_\speaker, \observation_\listener, \referents, \targetindex) &= \mathbbm{1}_{\targetindex = \hat{\targetindex}} \; .
\end{align*}
\end{small}
\vspace{-2.5em}

\subsection{Scene Generation}
\label{sec:scenegen}
Formally, we denote a scene $\scene = (\environment, \pose_\speaker, \pose_\listener, \referents, \targetindex)$ as an environment $\environment \in \environments$ populated with two agents $\pose_{\speaker}$ and $\pose_{\listener}$ and $N$ referents $\referents$, as well as the index of the target referent $\referent_\targetindex$.
To generate a scene, we first sample a base environment, then place the two agents, then the candidate referents. 
Finally, we render each agent's observation of the scene.\footnote{Appendix~\ref{app:scenegen} contains additional details about scene generation, including object placement and observation rendering.}

\paragraph{Base environments.}
We load indoor 3D environments from ScanNet++~\citep{yeshwanthliu2023scannetpp} as 3D meshes into habitat-sim~\cite{savva2019habitat}, which supports basic object physics and ray casting for identifying fields of view visible to each agent.

\paragraph{Agent placement.}
Both the speaker and listener agents are associated with a camera pose $\pose = (\langle x, y, z \rangle , \langle \theta, \phi, \psi \rangle)$, where $\langle x, y, z \rangle$ denote the position in 3D space and $\langle\theta, \phi, \psi\rangle$ represent the pitch, roll, and yaw angles respectively. 
To ensure observations are reasonable, we sample the camera height $z$ from a range of typical adult human height, and fix pitch $\theta$ and roll $\phi$ at 0$^\circ$.
We enforce a maximum distance between the agent cameras, and a non-empty overlap of their respective fields of view.
We randomly assign speaker and listener roles to the two cameras, except in the case that only one agent's camera is in the other's field of view, but not vice versa.
In this case, the former camera represents the speaker.

\paragraph{Candidate referent placement.}
Each scene contains a set of $N=3$ candidate referents $\referents = \left\{ \referent_1, \dots, \referent_N\right\}$, where $\referent_i = \langle x_i, y_i, z_i \rangle$ denotes the location of each referent. 
A target index $1 \leq \targetindex \leq N$ denotes the referent that the speaker aims to communicate to the listener.
For each referent, we first sample a position  from the set of all empty coordinates $\coordinates$ in the scene.
We use a gravitational physics simulation to drop the each referent from this position until it comes to rest on a solid horizontal surface.
We use rejection sampling to ensure all referents are visible to both agents, and referents are not too close together.

\paragraph{Agent observations.}
Each agent's observation is represented as a 2D image $\observation \in \reals^{3 \times H \times W}$ rendered from its camera pose $\rho$.
The speaker's observation $\observation_{\speaker} = \text{proj}_{\speaker}(\environment, \referents, \targetindex, \pose_{\speaker})$ is a projection of the speaker's view of the environment, and $\observation_{\listener} = \text{proj}_{\listener}(\environment, \referents, \pose_{\listener})$ is a projection of the listener's view.
The camera field of view is 90$^\circ$ both vertically and horizontally.
While $\text{proj}_{\listener}$ renders each referent with the same color (red), $\text{proj}_{\speaker}$ renders the target $\referent_{\targetindex}$ in a different color (blue) from the distractor objects, allowing the speaker to easily distinguish the target when writing their referring expression.
Both projections also render the other agent's camera as a 3D model of a human, which are sampled from 2K2K~\cite{han2023high}.

\subsection{Controlled Difficulty}\label{sec:difficulty}
We implement two ways to control the difficulty of referential communication via scene generation: by manipulating the relative orientation of speaker and listener, and by adversarially placing referents.
Figure~\ref{fig:view_angle} shows examples of four scenes generated from different relative orientations, and with and without adversarial referent placement.

\paragraph{Speaker-listener orientation.}
The relative orientation of the speaker $\pose_{\speaker}$ and listener $\pose_{\listener}$ is the absolute difference $\psi' = \min(|\psi_{\speaker} - \psi_{\listener}|, 360^{\circ} - |\psi_{\speaker} - \psi_{\listener}|)$ of their horizontal rotations (yaw). 
We experiment with the influence of $\psi'$ on interaction dynamics.
When $\psi'$ is close to $0^{\circ}$, the two agents are facing the same direction, and their observations are likely to be similar to one another.
When $\psi'$ is close to $180^{\circ}$, the agents are facing each other and thus have completely different views of the same scene.
Following \citet{schober1993spatial}, we hypothesize that differences in relative angles of speakers and listeners may influence language use.
Our environment supports uniformly sampling agent placements with fixed relative orientation.

\paragraph{Adversarial placement of referents.}
We design a referent placement policy model $\referentpolicy : \coordinates^*  \times \observations_{\speaker} \times \poses_{\speaker} \times \poses_{\listener} \rightarrow \Delta^{\referents^{N} \times \{ 1 \dots N\}}$, which takes as input a set of empty coordinates $\coordinates$, the speaker's observation prior to referent placement, and both agent poses.
It generates a distribution over referent locations prior to the physics simulation, and over referent indices representing the target. 
The policy model is implemented as a vision transformer~\cite{dosovitskiy2020image}, and is trained to maximize the communicative failure rate between two fixed agent models, $\hat{p}_\speaker$ and $\hat{p}_\listener$, by optimizing 

\vspace{-1em}
\begin{small}
\begin{equation*}
\max_{\referentpolicy} \mathbb{E}_{(\referents', \targetindex') \sim \referentpolicy(\cdot)} \left[ 1 - \text{Success}(\hat{p}_\speaker, \hat{p}_\listener, \observation_\speaker, \observation_\listener, \referents', \targetindex') \right] \; ,
\end{equation*}
\end{small}

\noindent where $\observation_\speaker$ and $\observation_\listener$ are the agents' observations after referents $\referents$ are placed.
During scene generation, we use the trained policy to sample initial positions of referents, then apply gravitational physics to find the resting position of each referent.

%% file: sections/05_experiments.tex
\section{Experimental Setup}\label{sec:agent}
We use our scene generation platform to evaluate embodied, multi-perspective referential communication with pairs of agents including humans and automated models.

\subsection{Data}\label{sec:data}
We generate a set of 27,504 scenes for training and evaluating automated agents.
We recruit crowdworkers to participate in the task both as listeners and speakers, collecting a dataset of 2,970 human-written referring expressions paired with human listener selections in 1,485 of these scenes.

\paragraph{Scene generation.}
We use ScanNet++~\cite{yeshwanthliu2023scannetpp} (non-commercial license), which contains 450 high-quality 3D indoor environments, as the basis of our task instances. 
We generate scenes using both forms of controlled difficulty (Section~\ref{sec:difficulty}).
First, we train our adversarial referent placement policy, implemented as ViT-s/16~\cite{dosovitskiy2020image}, using GPT-4o as both a speaker and listener in 27,600 generated scenes comprising 60 samples per base environment.\footnote{Appendix~\ref{app:dataadversary} contains more details on the adversary.}
To generate our final dataset of scenes, we first sample 300 agent placements for each relative angle in $\{0, \dots, \text{180}\}$ distributed uniformly across the 450 base environments.
For each of these agent placements, we sample two referent placements, resulting in two complete scenes: one where referent locations are randomly sampled, and another where referents are placed using the adversarial referent placement policy.

We use GPT-4o to perform rejection sampling on low-quality scenes. Our scene rejection process targets scenes where communication tasks become impossible or highly unrealistic. This includes scenes where referents are invisible to both parties, the image fidelity is extremely low, or referents defy physics by floating or clipping through the environment. 
We do not reject scenes that are simply difficult, e.g., due to object placement.
The final dataset includes 27,504 scenes, which we split into train (24,644 scenes), validation (1,485) and test (1,375) splits. The split is by scene instances. The validation split is used for ablating different dataset components or models, and the test split is to be used for testing final model performance.
Base environments may appear in multiple splits. 

\paragraph{Crowdsourcing.}
We recruit 194 crowdworkers on Prolific\footnote{\url{https://www.prolific.com}}.
Qualified workers are fluent English speakers, reside in the United States, and pass a qualification task by writing referring expressions for 15 scenes, with successful listener selection from two or more of three other workers for at least 10 of these referring expressions.
On average, we pay \$18 USD per hour.\footnote{Appendix~\ref{app:crowdsourcing} contains details on on data collection.}

\paragraph{Speaker task.}
Speakers are presented with a prompt that asks them to describe the location of the blue ball to another person who is always visible to them in the scene, and who cannot distinguish the colors of the balls. 
We make the listener always visible to the speaker to allow them to take into account the listener's perspective of the scene when writing a referring expression.
Speakers first click a button that reveals their view of the scene.
They write a referring expression, then submit their work.
We record both the referring expression and the time taken between revealing the scene and submitting the task.

\paragraph{Listener task.}
Listeners first click a button that reveals their view of the scene and a referring expression. 
They click on the referent they believe to be the target in the image, then submit their work. 
We record both the click position and the time taken between revealing the view and submitting the task.
A listener's selection is the sphere which is rendered closest to their click position.

\paragraph{Dataset statistics.}
For a randomly-sampled subset of 1,485 scenes from the validation set, we collect a referring expression from at least one worker, resulting in a total of 2,970 referring expressions, paired with judgments from three separate listeners.
Each referring expression is labeled with the majority-class referent selection.
The median time spent per speaker and listener task are 33.0s and 10.5s respectively. For all scenes, the speaker can see the listener; for 26\% of scenes, the listener can see the speaker.

\subsection{Evaluated Models}
\label{sec:eval_settings}

We experiment with four instruction-tuned vision-language models.\footnote{Additional details, including prompts, are available in Appendix~\ref{app:expsetup}.}
Two of these models are designed for more general use: \textbf{GPT-4o}\footnote{\url{https://openai.com/index/hello-gpt-4o/}}, a proprietary model developed by OpenAI that supports real-time joint processing of audio, vision, and text; and \textbf{LLaVA-1.5}~\cite{liu2023improvedllava}, a large open-weight instruction-tuned multimodal model.
We also experiment with two instruction-tuned open-weight models designed specifically to refer to regions of and ground references in images at any granularity: \textbf{Ferret}~\cite{you2023ferret} and \textbf{Groma}~\cite{ma2024groma}. 
Ferret employs a hybrid region representation that combines discrete coordinates and continuous features to represent regions in an image, while Groma utilizes a localized visual tokenization mechanism, where an image is decomposed into regions of interest and encoded into region tokens. 
We use these models as listeners only as preliminary experiments showed poor performance on reference generation.

\definecolor{CBBlue}{HTML}{0072B2} 
\definecolor{CBOrange}{HTML}{D55E00} 
\definecolor{CBGreen}{HTML}{009E73}
\newcommand\ra[2]{{#1} & {#2}}
\newcommand\raha[2]{\textcolor{CBOrange}{#1} & \textcolor{CBOrange}{#2}}
\newcommand\raah[2]{\textcolor{CBGreen}{#1} & \textcolor{CBGreen}{#2}}
\newcommand\reshead[1]{\multicolumn{2}{c}{\textbf{#1}}}

\begin{table*}[t!]
\centering\footnotesize
\begin{tabular}{rrcccccccccccc}
\toprule
& & \multicolumn{12}{c}{\textit{\textbf{Listeners}}} \\
\cmidrule(lr){3-14}
&  & \reshead{Human} & \reshead{GPT-4o} & \reshead{LLaVA-1.5} & \reshead{Ferret} & \reshead{Groma} & \reshead{ViperGPT} \\
& & Ran. & Adv.  & Ran. & Adv.  & Ran. & Adv.  & Ran. & Adv.  & Ran. & Adv.  & Ran. & Adv. \\
\midrule
\multirow{3}{*}{\rotatebox[origin=c]{0}{\textbf{\textit{Speakers}}}}  
& \textbf{Human}       & \ra{\textbf{\textcolor{CBBlue}{90.2}}}{\textbf{\textcolor{CBBlue}{84.9}}}  & \raha{67.6}{66.0} & \raha{63.3}{63.2} & \raha{70.1}{68.2} & \raha{64.3}{65.7} & \raha{57.8}{56.0} \\ 
& \textbf{GPT-4o} & \raah{67.8}{62.0} & \ra{61.1}{57.2} & \ra{60.4}{57.8} & \ra{67.8}{62.1} & \ra{66.5}{64.8} & \ra{55.6}{53.3} \\ 
& \textbf{LLaVA-1.5} & \raah{55.2}{56.1} & \ra{50.9}{49.8} & \ra{44.7}{42.2} & \ra{59.1}{52.8} & \ra{61.9}{55.4} & \ra{48.9}{48.7}  \\ 
\bottomrule
\end{tabular}
\caption{Rates of communicative success for all four combinations of human and automated speakers and listeners, across 1,485 scenes, split by scenes with random (Ran.) and adversarial (Adv.) referent placement. 
Results for human-human pairs are bolded and in \textcolor{CBBlue}{\textbf{blue}}; results for human speakers and automated listeners are in \textcolor{CBOrange}{orange}; results for human listeners and automated speakers are in \textcolor{CBGreen}{green}; and results for fully-automated pairs are in black.
}
\label{tab:results}
\vspace{-10px}
\end{table*}

We also experiment with modular vision-language reasoning systems, which decompose the problems of language understanding and perception by first mapping language to some executable code, which is then executed on an image~\cite{subramanian2023modular,gupta2023visual}.
In this work, we use \textbf{ViperGPT}~\cite{surismenon2023vipergpt}, using GPT-4 to generate intermediate Python programs. 
We use ViperGPT as a listener agent only.

For both speaker models, we provide as input the speaker's observation $\observation_\speaker$ and a prompt to describe the location of the blue sphere. 
For listeners, we provide as input a referring expression $\refex$ and the listener's observation $\observation_\listener$, as well as a list of each candidate referent's bounding box, and prompt the model to select the bounding box corresponding to the described target.
We sample from all models using a temperature of 0.

\subsection{Evaluation and Analysis}
We evaluate models both as speakers and listeners, partnered both with human and automated agents.
Our main metric is communicative success: for each scene, did the pair of agents successfully coordinate on the target referent?
Pairing automated listeners with human speakers evaluates a model's ability to comprehend a human-written referring expression, and pairing automated speakers with human listeners evaluates a model's ability to precisely refer to a region of the scene.
Both sides of this communicative task require understanding spatial language and taking into account the other agent's perspective of the shared scene.
For each setting, we analyze the influence of task difficulty on communicative success.

%% file: sections/056_human_study.tex
\section{Results}
\label{sec:eval}

We experiment with four configurations of agent dyads, combining humans and automated speakers and listeners.
Table~\ref{tab:results} includes results for the 1,485 validation scenes we use for collecting human-human data, split across scenes with random and adversarial referent placement.

\paragraph{Human speakers and listeners.}
Using the referring expressions collected in Section~\ref{sec:data}, we find that human-human pairs achieve an average communicative success rate of 87.6.\footnote{For fair comparison to settings where only one referring expression is produced per scene, we report the macro-average over scenes. The micro-average over all referring expressions in this experiment is 88.4.}

\paragraph{Human speakers, automated listeners.}
We evaluate model performance in comprehending human-written referring expressions. 
For each human-written referring expression in our collected dataset, we select the most likely referent according to the model.
We observe substantially lower accuracy in referent selection compared to human listeners.
Ferret, which was designed for fine-grained vision-and-language processing, outperforms the other models at an average selection accuracy of 69.2, but still lags far behind human performance.

\paragraph{Automated speakers, human listeners.}
We acquire a single referring expression from each instruction-tuned model for each evaluation scene.
For each referring expression, we acquire three human listener selections and compare the majority class referent to the intended target.
Both GPT-4o and LLaVA-1.5 are significantly less successful in describing target referents when compared to human speakers; GPT-4o's references lead to correct human listener selection in 64.9\% of scenes, while the LLaVA-1.5 speaker is successful for 55.7\%.

\begin{figure*}[t]
\centering
\includegraphics[width=0.99\textwidth, trim={0 0 0 0}, clip]{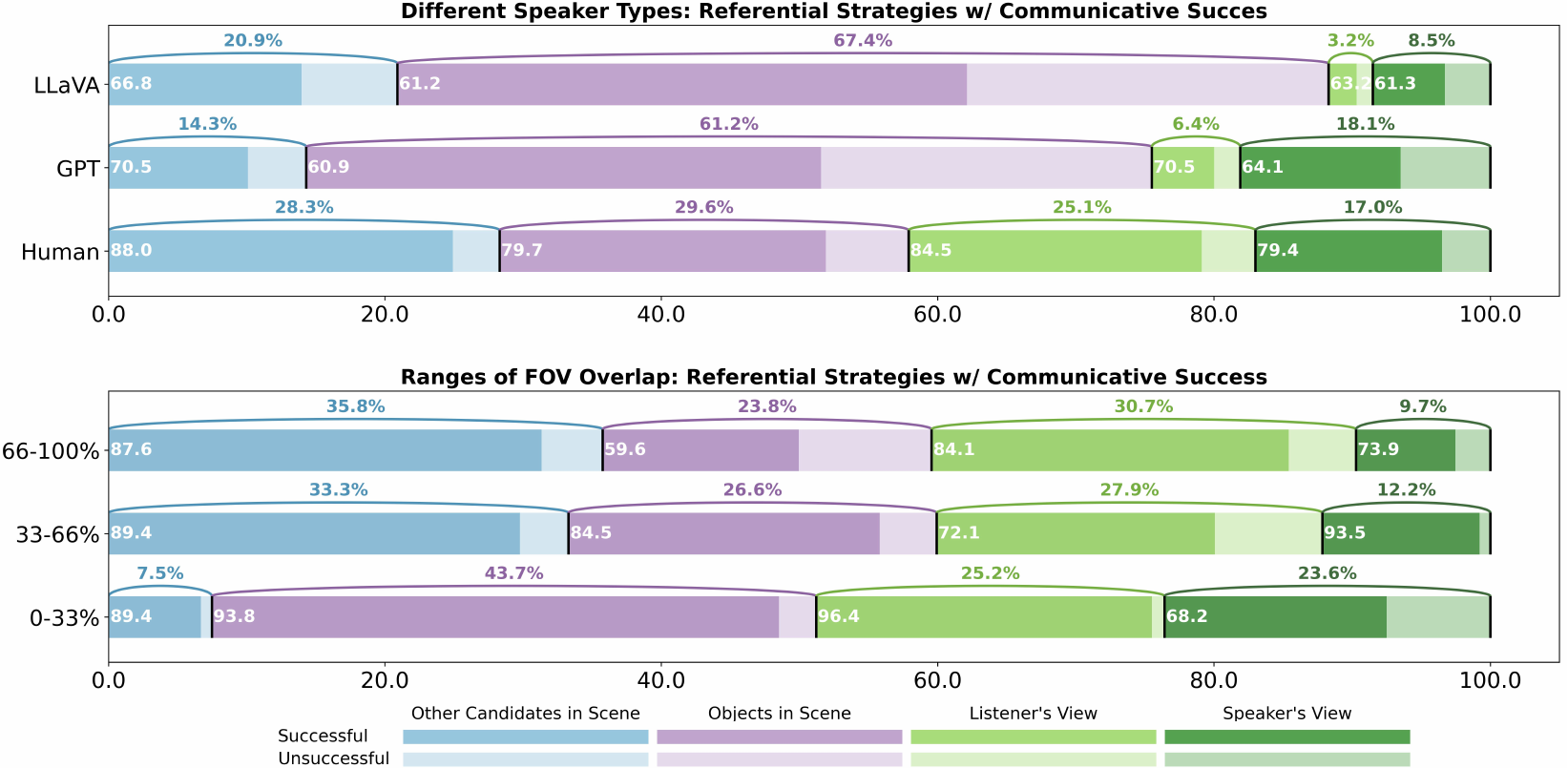}
  \caption{Analysis of referential strategies with respect to speaker agent type (top) and ranges of overlap in field of view (bottom).
  For each speaker agent or range of overlap, we plot the distribution over four referential strategies across all validation scenes. 
  Within each referential strategy, we also report the proportion of generated references that guide a human listener to successfully select the target reference.
  }\label{fig:reference_results}
\vspace{-10px}
\end{figure*}

\paragraph{Automated speakers and listeners.}
We evaluate settings where both agents are automated systems.
Using the referring expressions acquired from both speaker agents, we use all five listener models to perform referent selection.
In nearly all cases, performance with pairs of automated listeners is lower than dyads containing at least one human.
However, both Ferret and Groma perform on par with human listeners on referring expressions generated by both GPT-4o and LLaVA-1.5, for both random and adversarial referent configurations.
In fact, both models actually outperform human listeners for referring expressions generated by LLaVA-1.5 for random referent configurations.

\subsection{Adversarial Referent Placement}
Our adversarial referent placement policy was trained to minimize communicative success between a GPT-4o speaker and listener.
Table~\ref{tab:results} shows that scenes generated with this policy indeed reduce rates of communicative success in this setting by 3.9\%, a statistically significant difference confirmed by a paired t-test ($p <$ 0.05).
The learned policy also reduces the success rate for nearly all other combinations of agents, including for human-human pairs, where we see rates of communicative success drops from 91.6 to 85.1 when adversarially placing candidate referents.

\subsection{Language Analysis}

We manually annotate 200 randomly-sampled referring expressions written by crowdworkers and GPT-4o with respect to referential strategies used by the speaker. 
Then, to scale to all validation data, we use GPT-4o to categorize referential strategy given in-context examples selected from these 200 examples.
We consider four core referential strategies: reference to other candidate referents (e.g., \emph{in front of the other two red balls}), reference to fixed objects in the scene (\emph{in front of the kitchen entryway}), and reference to the listener (\emph{on your left}) or speaker's perspective (\emph{closest to me}).

Figure~\ref{fig:reference_results} (top) shows the prevalence of each referential strategy for both human and automated speakers in the validation set. 
Overall, our analysis shows that, compared to humans, automated models are more likely to refer to the target's relative position among objects in the scene, and much less frequently refer to its position with respect to the listener's view.
This policy is detrimental to model performance:
LLaVA especially fails to \textit{accurately} refer to other objects in the scene when describing the target, with only 61.2\% of such references resulting in communicative success.

We also analyze the influence of view similarity between both agents on referential strategies and communicative success (Figure~\ref{fig:reference_results}, bottom).
We compute field of view overlap\footnote{Field of view overlap is computed as the intersection over union of both agent's view on each candidate referent's surface. For example, if the speaker sees the front of a sphere and the listener is positioned to see the back of it, the overlap will be very low. Overlap is averaged over all candidate referents.} as a proxy for view similarity.
As the speaker's observations become increasingly similar to the listener's, they tend to describe the target with respect to other candidate referents.
As their views become dissimilar, speakers shift strategies to refer to targets with respect to other objects in the scene, and with respect to their own perspective~\cite{schober1993spatial}.

%% file: sections/055_learning.tex
\section{Learning from Communicative Success}
\label{sec:commsuccess}
We propose to further train our speaker model from learning signals acquired during referential communication.
The basic premise that motivates this approach is that empirical observations of language interpretation provides evidence of utterance meaning, regardless of speaker intent~\cite{kojima2021continual}.
For instance, if the listener selects a different referent than the intended target, this indicates the speaker's referring expression describes (or at the very least, better describes) the chosen referent, even if the generated expression fails to describe the intended referent.
In contrast to prior work that proposes methods that learn from communicative success (or failure)~\cite{kojima2021continual,liu2023computational}, we additionally explore the use of  preference-based learning signals that explicitly pair the intended and chosen targets in case of communicative failure.

\paragraph{Learning.}
During training, we collect a dataset of $M$ examples 
$\dataset = \left\{ (\scene^{(i)}, \refex^{(i)}, \hat{\targetindex}^{(i)}) \right\}^M_{i=1}$, 
each consisting of a generated scene $\scene$ (including the target referent index $\targetindex$), referring expression $\refex \sim p_\speaker(\observation_{\speaker}, \referents, \targetindex; \theta)$ sampled from a pre-trained speaker  and the referent $\hat{\targetindex} \sim p_\listener(\observation_{\listener}, \referents, \refex; \phi)$ selected by a listener.
We fine-tune speaker parameters $\theta$ using our collected dataset of examples $\dataset$.

We experiment with four methods for using the collected data: (a) contrastive learning~\cite{radford2021learning}, (b) learning from successes only (LSO), (c) creating positive examples from every example (Pos. Only), and (d) pairwise preference learning (PPL).
In contrastive learning, for examples where $t \neq \hat{t}$, we apply a contrastive objective to jointly maximize the probability of $x$ given the chosen referent $\hat{t}$ and minimize the probability of $x$ given the intended referent $t$.
For all other methods, we use offline proximal policy optimization~\citep[PPO; ][]{schulman2017proximal}, adjusting only the reward function.
When learning from successes only, examples receive a reward of +1 when $\targetindex = \hat{\targetindex}$ and 0 otherwise.
To create positive examples from every example, we assign a positive reward of +1 to each utterance $x$ paired with the listener's selection $\hat{t}$, which may or may not be equivalent to $t$.
In pairwise preference learning, we take advantage of the fact that, especially in light of communicative failure, we can assume that the referring expression better describes the listener's guess than the speaker's intended referent.  We formalize this with a reward function that maximizes the difference between the likelihoods of the speaker's referring expression $\refex$ describing the listener's chosen target $\hat{\targetindex}$ versus the intended target $\targetindex$:

\begin{small}
    \begin{equation*}
  p_\speaker (\refex \mid \observation_\speaker, \referents, \hat{\targetindex}; \theta') - p_\speaker(\refex \mid \observation_\speaker, \referents, \targetindex; \theta') \; . 
    \end{equation*}
\end{small}

\noindent In cases where $\targetindex = \hat{\targetindex}$, the assigned reward is +1.

Finally, we also experiment with imitation learning, where we acquire human-written references.
For each reference, we acquire three human listener selections.
For each selection, we directly fine-tune the speaker model parameters to maximize the probability of the human reference conditioned on the scene and listener selection. 

\paragraph{Experimental setup.}
We use the initial speaker model, pre-trained LLaVA-1.5~\cite{liu2023improvedllava}, to generate referring expressions for 200 scenes sampled from the training split.
We experiment with learning from both human and automated listener agents.
We hypothesize that human listeners will provide higher-quality feedback in the form of referent selections than the automated listener model, given a human listener's superior language-understanding capability.
However, using an automated listener is less costly, as it requires collecting no additional human data.
For our automated listener, we also use pre-trained LLaVA-1.5.
We collect a single guess per referring expression from our automated listener, and three human listener guesses.
This results in two datasets: $\dataset_a$ containing 200 examples of automated listener selections, and $\dataset_h$ containing 600 examples of human selections.
Both datasets contain the same 200 sampled speaker references.
Training results in eight models trained on model-generated references: for each of the training objectives (Contrastive, LSO, Pos. Only, and PPL), we learn from automated and human listener selections ($\dataset_a$ and $\dataset_h$).
For the same 200 scenes, we also acquire one human-written referring expression and 3 listener selections for imitation learning.

For evaluation, we acquire three human listener selections for generated referring expressions in a randomly-sampled but representative subset 195 scenes from the validation set. 
For the best-performing and baseline models, we also evaluate on our test set of 1,375 scenes.

\begin{table}[t]
\centering\footnotesize
\resizebox{\columnwidth}{!}{
\begin{tabular}{lccc}
\toprule
\textbf{Speaker} & \multicolumn{2}{c}{\textbf{Listener Accuracy}} & \textbf{Avg. Ref.}\\
& \textit{Val.} & \textit{Test} & \textbf{Length} \\
\midrule
Pre-trained $\theta$ & 59.7 & 58.9 & 61.1 \\
+ Contrastive ($\dataset_a$) & 60.9 & -- & 45.8 \\ 
+ Contrastive ($\dataset_h$) & 62.1 &  -- & 55.7 \\
+ LSO ($\dataset_a$) & 61.5 &  -- & 41.7 \\
+ LSO ($\dataset_h$) & 65.6 &  -- & 54.6 \\
+ Pos. Only ($\dataset_a$) & 62.1 &  -- & 46.7 \\
+ Pos. Only ($\dataset_h$) & 66.0 &  -- & 57.2\\
+ PPL ($\dataset_a$) & 66.7 &  -- & 19.8 \\
+ PPL ($\dataset_h$) & 69.2 & 69.3 & 15.6\\
+ Imitation Learning & 67.9 & 68.2 & 16.8 \\
\midrule
Human & 91.3 & 90.6 & 15.8  \\
GPT-4o & 66.3 & 67.1 & 78.9  \\
\bottomrule
\end{tabular}}
\caption{Performance of the LLaVA-1.5 speaker before and after training on data collected in 200 scenes with human and automated listeners, as well as performance of human and GPT-4o speakers on the same set of scenes.
We also report the average reference length for each speaker.
}
\vspace{-15pt}
\label{tab:learning_results}
\end{table}

\paragraph{Results.}
Table~\ref{tab:learning_results} shows that learning from communicative success significantly improves the quality of an initially-weak speaker agent. 
Overall, learning from human listeners ($\dataset_h$) is significantly more effective than learning from an automated listener, though this is still beneficial.
We also find that preference learning (PPL) significantly\footnote{Using a paired t-test, we find $p <$ 0.05 when comparing LSO and PPL for both fine-tuning dataset and $p <$ 0.05 when comparing Pos. Only and PPL.} improves over training only on examples exhibiting correct target selection.
After fine-tuning on only 200 sampled referring expressions with human judgments and preference-based reward, LLaVA-1.5 actually outperforms GPT-4o as a speaker, with a communicative success rate of 69.3 when paired with human listeners. 
This approach also performs comparatively to imitation learning, which is more costly due to requiring human-written references.

Manual analysis on the validation examples reveals that after training, the model generates fewer genuinely ambiguous descriptions (43.6 to 36.0\% of analyzed descriptions), and shifts from a referential strategy that increasingly refers to the listener (3.2 to 20.6\%) or speaker (8.5 to 21.3\%) perspectives.
We also analyze how training influences sentence length: prior to training, LLaVA-1.5 produces lengthy descriptions at an average length of 61.1 tokens.
For all training objectives, the fine-tuned model generates shorter expressions than the initial model.
However, only after applying PPL-based learning does the sentence length decrease close to lengths of human references, without training on any human references.

%% file: sections/03_related.tex
\section{Related Work}
The meanings of relative spatial terms are highly dependent on the situated environment: the items participating in the relation and their intrinsic parts and affordances~\citep{clark1973space,landau2018learning}; the relative perspectives of participants in an embodied scene~\citep{taylor1996perspective,goschler2008perspective}; and within-interaction conventions formed during multi-turn embodied dialogue~\citep{schober1993spatial}, among other factors.
In this work, we focus on the influence of relative perspective between multiple on the use of spatial language.

Production and comprehension of referring expressions has been studied in human-human dialogue \citep[][\textit{inter alia}]{clark1986referring,taylor1996perspective,vandersluis2011cross,udagawa2020linguistic}, and in interactions between human and automated language users~\citep[][\textit{inter alia}]{janarthanam2010adaptive,fang2014collaborative,fang2015embodied,huang2020generating}. 
However, most work has focused on disembodied referential communication, where agents tasked with communicating about sets of stimuli~\cite{hawkins2017convention,haber2019photobook}, or where agents are not physically situated within an environment~\cite{kazemzadeh2014referitgame,achlioptas2020referit3d}.
The problem of situated language grounding in multi-agent settings reflects an increasingly popular real-world scenario of embodied agents.
In studies where interaction participants are both embodied with different visual perspectives on the same scene, they must either be literally physically embodied in a single scene~\cite{schober1993spatial,taylor1996perspective}, or are placed in synthetic environments~\citep{Udagawa2019natural}.

A small number of existing works have trained language-generation models using evidence of communicative success in interaction with another agent.
For example, \citet{kojima2021continual} train an instruction-generating agent by observing humans follow generated instructions, and \citet{liu2023computational} use signals from reference games with automated listeners to improve a speaker model.
Our work takes inspiration from the latter to improve our speaker model using referent selections from an automated listener; however, we explore a preference-based objective that explicitly pairs the intended and empirically chosen referents.

%% file: sections/06_discussion.tex
\section{Conclusion}
We study multi-agent referential communication in situated interactions.
In this setting, a speaker and a listener are both embodied in a shared scene, but are placed in different locations, with different views of the scene.
We design a platform that supports generation of photorealistic 3D scenes, with control for difficulty of the referential task.
We evaluate both humans and automated agents as speakers and listeners in this task.
While human-human dyads are successful at coordinating on a referent around 88.4\% of the time, automated models fall far behind when used both as speakers and as listeners.
However, we can substantially improve the performance of an open-weight speaker model by training it with evidence of communicative success in referential communication with both automated and human listeners.
Our findings suggest that despite the increasing relevance of multi-agent situated interactions between humans and automated agents, there is significant headroom for applying models that jointly process language and visual perception in this setting.
However, they also show the promise of training such agents in interaction with people.

\section*{Limitations}
Our task currently focuses on single-shot reference, where a speaker creates a single referring expression, and the listener cannot ask for clarification or engage in interactive reference resolution~\cite{clark1986referring,Udagawa2019natural}.
Evaluating how models participate in an interactive version of our task is a compelling direction for future work.
Additionally, while our experiments are currently conducted exclusively in English, the language of space and motion has enormous variation across language communities~\citep{levinson2006grammars}.
Core spatial concepts studied in English, like \textit{on} or \textit{in}, do not have universally uniform meanings, with different languages dividing the conceptual space of spatial language in vastly different ways~\cite{landau2017update}.
Future work should explore how spatial concepts and referential strategies vary across movement and non-static environment, multi-turn conversations, language features, and more complex scenarios.
Finally, our experiments on learning from communicative success perform only a single round of speaker deployment and training.
Future work could perform further rounds of speaker deployment and listener judgments \cite[i.e., as in][]{kojima2021continual,suhr2023continual}, and analyze dynamics of language change in a continual learning setting.

\section*{Acknowledgments}
This work was supported by a Young Investigator Grant from the Allen Institute for AI. We thank the Berkeley NLP group and the anonymous reviews for their advice and suggestions on our work.

%% file: sections/appendix.tex
\clearpage
\definecolor{CBBlue}{HTML}{0072B2} 
\definecolor{CBOrange}{HTML}{D55E00} 
\definecolor{CBGreen}{HTML}{009E73}

\section{Data}

\subsection{Scene Generation}\label{app:scenegen}

\paragraph{Agent placement.}
We impose three constraints on agent placement to help a more efficient scene generation pipeline:
\begin{itemize}
    \item Maximum distance between the agents: Let $d_{\max}$ be the maximum allowed distance between the speaker and the listener. Denoting the positions of the speaker and listener as $\pose_{\speaker}$ and $\pose_{\listener}$, respectively, we require that $|\pose_{\speaker} - \pose_{\listener}| \leq d_{\max}$. We use $d_{\max}=10$.
    \item Field of view overlap: Let $\fov_{\speaker}$ and $\fov_{\listener}$ be the fields of view of the speaker and listener, respectively. We require that the intersection of their fields of view is non-empty, i.e., $\fov_{\speaker} \cap \fov_{\listener} \neq \emptyset$.
    \item Relative viewing angle: Let $\psi_{\speaker}$ and $\psi_{\listener}$ be the horizontal viewing angles of the speaker and listener, respectively, relative to a common reference direction. The relative viewing angle between the agents is given by $\psi' = \min(|\psi_{\speaker} - \psi_{\listener}|, 360^{\circ} - |\psi_{\speaker} - \psi_{\listener}|)$. We can place the agents with a pre-set relative viewing angle by satisfying $C_0 \leq |\psi'_{\speaker} - \psi'_{\listener}| \leq C_1$, where $C_0$, $C_1$ is the viewing angle difference bounds we set.
\end{itemize}

\paragraph{Referent placement.}
We impose three constraints on referents placement so they don't stack, become obstructed, or float in the air to meet real world physics standards:
\begin{itemize}
    \item Visibility constraint: Let $\visibleset_{\speaker}$ and $\visibleset_{\listener}$ be the sets of points visible from the speaker's and listener's cameras, respectively. For each referent $\referent_i$, we require that $\referent_i \in \visibleset_{\speaker} \cap \visibleset_{\listener}$.
    \item Physically-based placement: Let $\mathcal{X}, \mathcal{Y}, \mathcal{Z}$ be the sets of valid $x$, $y$, and $z$ coordinates within the environment bounds. For each referent $\referent_i$, we randomly sample coordinates $(x_i, y_i, z_i) \in \mathcal{X} \times \mathcal{Y} \times \mathcal{Z}$ and drop the referent using gravitational physical simulation until it comes to rest on a solid horizontal surface.
    \item Minimum distance: Let $d_{\min}$ be the minimum required distance between any two referents. For all pairs of referents $\referent_i$ and $\referent_j$, where $i \neq j$, we enforce $|\referent_i - \referent_j| \geq d_{\min}$. We use $d_{\min}=0.3$ .
\end{itemize}

\paragraph{Scene rendering.}
Our environment supports rendering observations at different resolutions; e.g., we use $H$ = 720 and $W$ = 1280 for HD resolution. 
For environment generation, we use Quadro RTX 6000 for graphics rendering for a single process. We parallelize data generation with Habitat-Sim with 4 Quadro RTX 6000.

\paragraph{Scene rejection sampling.}
We use GPT-4v to discard low quality images rendering during the dataset generation. We use the following prompt:\\

\noindent\fbox{%
    \parbox{0.95\columnwidth}{%

\begin{small}
\texttt{
Please analyze the following image and provide a score from 0 to 10 based on these criteria:
\begin{itemize}
    \item The image must contain exactly 3 red spheres. If there are more or fewer than 3 red spheres, the score should be 0.
    \item The image should have high perceptual quality. Consider factors such as:
    \begin{itemize}
        \item \textbf{Resolution}: The image should be clear and not pixelated or blurry.
        \item \textbf{Lighting}: The image should have adequate lighting, without extreme darkness or overexposure.
        \item \textbf{Focus}: The subject of the image (the red spheres) should be in focus.
        \item \textbf{Contrast}: The image should have good contrast, allowing the red spheres to be easily distinguishable from the background.
    \end{itemize}
    \item The image should not have any visible artifacts, such as:
    \begin{itemize}
        \item \textbf{Compression artifacts}: There should be no visible compression artifacts, such as blocky patterns or color banding.
        \item \textbf{Noise}: The image should not have excessive noise or graininess.
        \item \textbf{Distortions}: The image should not have any distortions, such as warping or stretching.
    \end{itemize}
\end{itemize}
}
\end{small}
    }%
}

\subsection{Adversarial Referent Placement}\label{app:dataadversary}
For each training iteration, the vision transformer (ViT-s/16) takes as input the speaker view, and the available object placement locations and speaker and listener locations processed as $(x, y, z)$ coordinates flattened into a normalized array. The model is trained to output the hard location from the input object placement locations as a single-choice pipeline. 

\subsection{Crowdsourcing}\label{app:crowdsourcing}
For speakers and listeners we prompt the user to follow a description and a tutorial. When annotating, they still have access to the tutorial.
They are provided the following task description:\\

\vspace{-13.8px}
\noindent\fbox{%
    \parbox{0.95\columnwidth}{%

\begin{small}
\texttt{
We engage participants in a virtual environment where they assume the roles of a Speaker and a Listener. The task involves communication and spatial reasoning, requiring the ``Speaker'' to describe the location of specific objects within the environment, which are visible to them but not to the Listener. The Listener then interprets these descriptions to identify the objects accurately. Data collected from these interactions helps us understand the effectiveness of communication strategies and spatial language in varied settings. This study aims to improve collaborative tasks between humans and AI agents, enhancing how they interact within real-world environments.}
\end{small}
}}
\vspace{2pt}

We qualify participants from the USA who are fluent in English. Users are informed their data will be used for research purposes. Our study is determined exempt from UC Berkeley CPHS.
We manually check human data for non-conforming text. This step includes excluding private user information or offensive content.

\section{Experiments}

\subsection{Experimental Setup}\label{app:expsetup}

We prompt the instruction-tuned vision and language models to output speaker and listener text. Except for the model-specific architecture input formatting. We use the following prompts:\\

\vspace{-5pt}
\noindent Speaker Prompt: \\
\noindent\fbox{%
    \parbox{0.95\columnwidth}{%
\begin{small}
\texttt{
Describe the location of the blue sphere relative to the environment features, relative to your view and the other person's view, and in contrast with other red spheres.
}
\end{small}}}
\vspace{2pt}

\noindent Listener Prompt:\\

\noindent\fbox{%
    \parbox{0.95\columnwidth}{%

\begin{small}
\texttt{An image filled with several identical red spheres and a blue sphere. Your task is to identify the specific red sphere of interest from among several possible candidates. To assist you, you will receive a detailed description highlighting unique characteristics or positions of the sphere.}

\texttt{Your objective is to determine the precise location of this sphere in the image and mark it with a bounding box. Consider factors such as lighting, reflections, shadows, relative position to other objects, and any unique attributes mentioned in the description. You should analyze how these details help to pinpoint the exact sphere among the identical ones.}
\end{small}}}
\noindent\fbox{%
    \parbox{0.95\columnwidth}{%
\begin{small}
\texttt{Once you have identified the sphere, outline its position using a bounding box and provide its coordinates in the format:}

\texttt{$x_0$ (left), $y_0$ (top), $x_1$ (right), $y_1$ (bottom)}

\texttt{Additionally, explain your reasoning in detail for why you chose this specific location for the bounding box. For example:}

\texttt{``Based on the description, the sphere is near the window on the left side, and the distinct light reflection on its surface sets it apart from the others. This suggests its location as... , Bounding box coordinates: [0.23, 0.44, 0.30, 0.46].''}

\texttt{Be aware that the description might offer a different viewpoint of the scene, so be prepared to adjust your analysis accordingly.}

\texttt{\textbf{Choose from the following bounding boxes:} [candidate bounding boxes]}

\texttt{\textbf{Format for Response:}}

\texttt{\textbf{Reasoning for location choice:} [Your detailed explanation here]}

\texttt{\textbf{Bounding box coordinates:} [$x_0$, $y_0$, $x_1$, $y_1$]}

\texttt{Feel free to incorporate any nuanced observations or contrasting elements that helped you make the distinction.}

\end{small}}}

\subsection{Influence of Speaker Visibility}
\begin{table}[h]
\centering\footnotesize
\resizebox{0.999\columnwidth}{!}{
\begin{tabular}{rrcccc}
\toprule
& & \multicolumn{4}{c}{\texttt{\textbf{Listeners}}} \\
\cmidrule(lr){3-6}
&  & \reshead{\textbf{Human}} & \reshead{\textbf{GPT-4o}} \\
& & Visible & Not Visible & Visible & Not Visible \\
\midrule
\multirow{2}{*}{\textbf{\texttt{Speakers}}}  
& \textbf{Human}       & {87.5} & {86.1}  & {67.2} & {66.0} \\ 
& \textbf{GPT-4o}      & {65.8} & {65.4} & {60.4} & {59.2} \\ 
\bottomrule
\end{tabular}}
\caption{Influence of speaker visibility to listener on listener performance.}
\label{tab:performance}
\end{table}

In 26\% of generated scenes, the speaker is visible to the listener agent. 
We find that for human speakers, the visibility of the speaker significantly (though only slightly) increases  communicative success ($p < $ 0.01 using a paired t-test), while the difference is not significant for GPT-4o based speakers.

\begin{figure*}
\centering
\includegraphics[width=0.99\textwidth]{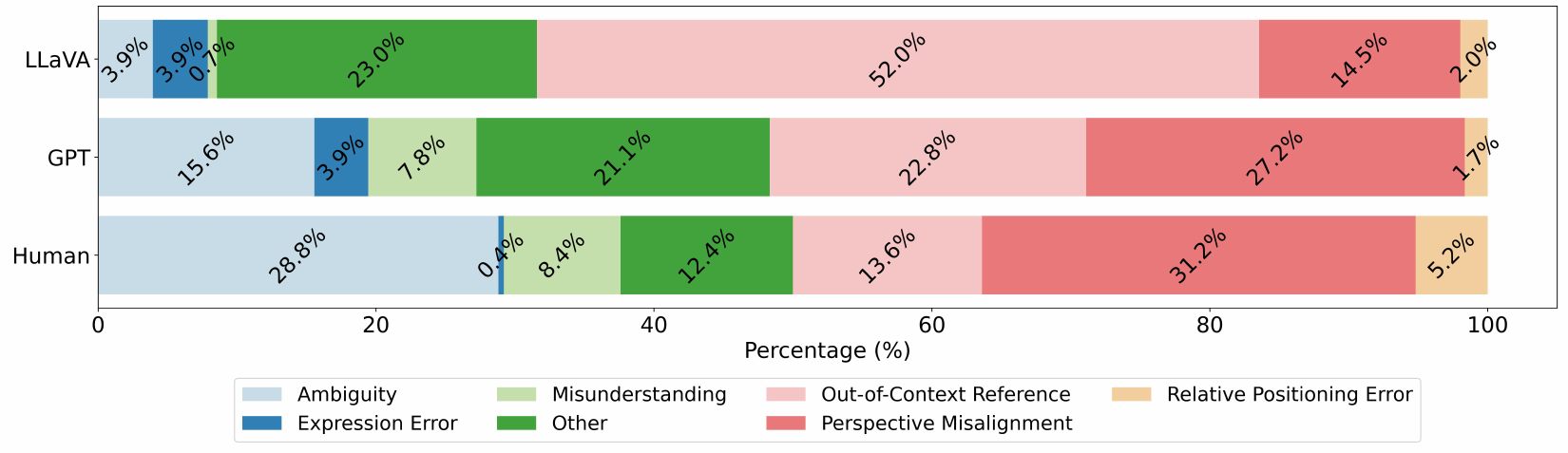}
  \caption{Impact of task difficulty on communication errors between speaker and listener for Human, GPT, LLaVA speakers.
  }\label{fig:error_distribution}
\end{figure*}

\subsection{Error Example}
\begin{figure}[h]
\centering
\includegraphics[width=0.999\columnwidth]{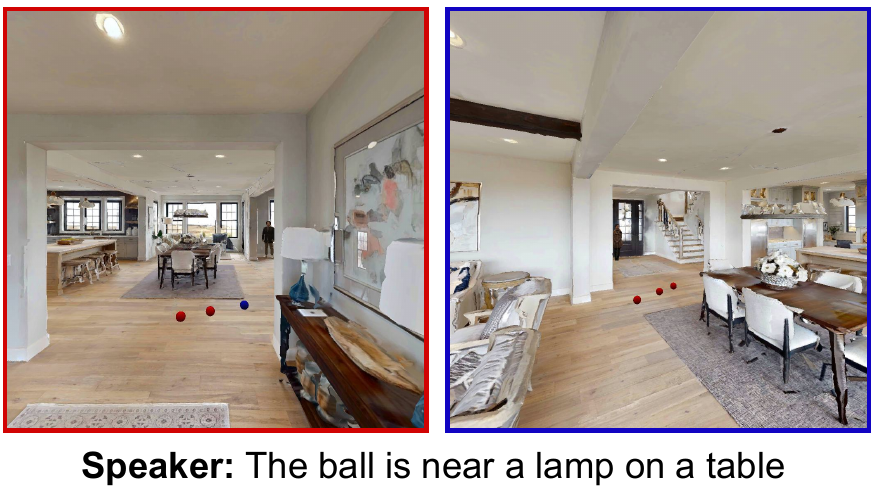}
  \caption{LLaVA speaker example that leads to incorrect listener selection.
  }\label{fig:error_example}
\end{figure}

We analyze the frequency of several common communication errors in collaborative tasks involving both human and automated speakers interacting with human listeners, with varying degrees of task difficulty. Out-of-context reference is when speaker reference context that is not in listener's view;. Perspective misalignment is when speaker reference its own perspective which will change drammatically when switched to listener's perspective. Ambiguity is that speaker expression can resolve to different meanings according to views. Relative position error is when the speaker expression describes wrong relative position like 'to the left of'. Expression error is simply wrong expression. Misunderstanding is when the speaker expression is unambiguously correct but listener fails to resolve it. The results are presented in Fig~\ref{fig:error_distribution}. It is evident that the error frequency in collaborations involving LLaVA speakers is generally higher than other speakers. Most errors are predominantly out-of-context reference, perspective misalignment, and ambiguity.
For example, in Figure \ref{fig:error_example}, LLaVA mistakenly reference objects that are not in the view of the listener.

The impact of facing angles and distances on communication is also significant. We find that errors are most prevalent when the listener and speaker are facing each other at angles between 120-180 degrees. 
In these situations, directional terms such as ``left'' and ``right'' often become inverted, especially when speakers fail to clarify whose perspective is being used. Moreover, with the visibility of both parties, a speaker might use ``human'' as a reference point, but the listener typically assumes ``human'' refers to the speaker, leading to selections in the opposite direction. Additionally, as the distance between speaker and listener increases, the descriptions provided by speakers tend to become more vague, opting for broader reference points such as ``on the left side of the wall'' rather than ``next to the table'', further complicating accurate communication.

\subsection{View Overlap Analysis}
\begin{figure}[h]
\centering
\includegraphics[width=0.999\columnwidth]{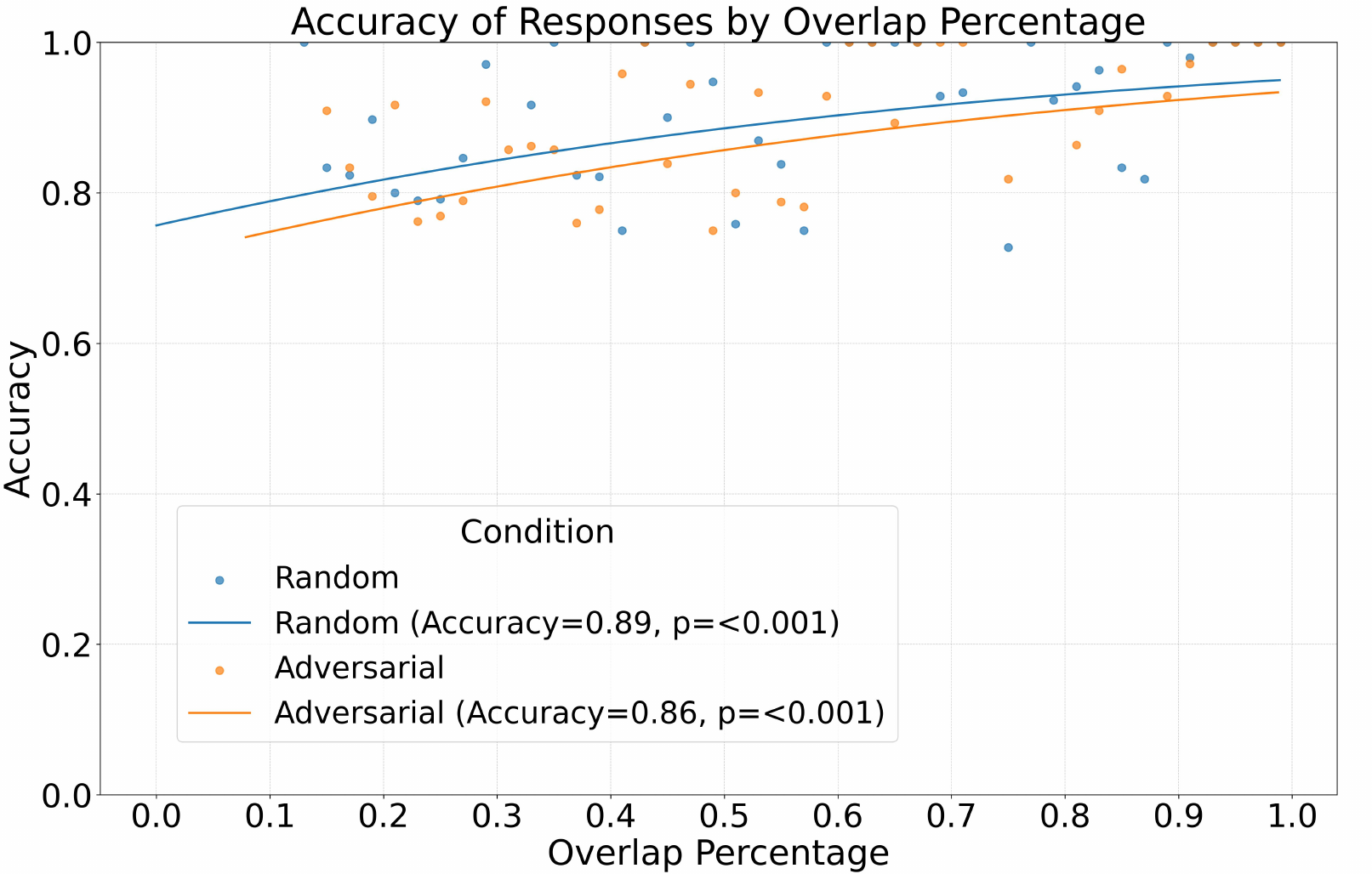}
  \caption{Overlap of object and distribution of correct listener selection.
  }\label{fig:view_overlap_analysis}
\end{figure}
We perform analysis on speaker and listener view overlap, which is calculated by the percentage of objects area seen by speaker and listener. We use logistic regression on individual data points with likelihood ratio test (LRT) both p-values<0.001. And we calculate accuracy over 0.02 interval of buckets on the overlap percentage for the scatter plot and Chi-Square test with p-value<0.05. Higher overlap usually means speaker and listener have close view pose and position. We can see from the plot that for both adversarial and random placements, as the view overlap increases, the performance is better. 

\subsection{AI Assistants Usage}\label{app:aiuse}
When conducting this research, we use AI to enchance our coding efficiency and quality. We use ChatGPT \footnote{https://chat.openai.com/} and Claude.ai\footnote{https://claude.ai} to assist in writing code for dataset generation and the human study website server.